\documentclass{article}

\usepackage[final]{corl_2023} 
\usepackage{graphicx}

\title{Object Recognition and Force Estimation with the GelSight Baby Fin Ray}

\author{
  Sandra Q. Liu\\
  Department of Mechanical Engineering\\
  Massachusetts Institute of Technology
  United States\\
  \texttt{sqliu@mit.edu} \\
  \And
  Yuxiang Ma\\
  Department of Mechanical Engineering\\
  Massachusetts Institute of Technology
  United States\\
  \texttt{yxma@csail.mit.edu} \\
  \And
  Edward H. Adelson\\
  Department of Brain and Cognitive Sciences\\
  Massachusetts Institute of Technology
  United States\\
  \texttt{adelson@csail.mit.edu} \\
}

\begin{document}
\maketitle


\begin{abstract}

Recent advances in soft robotic hands and tactile sensing have enabled both to perform an increasing number of complex tasks with the aid of machine learning. In particular, we presented the GelSight Baby Fin Ray in our previous work\cite{bbfinray}, which integrates a camera with a soft, compliant Fin Ray structure. Camera-based tactile sensing gives the GelSight Baby Fin Ray the ability to capture rich contact information like forces, object geometries, and textures. Moreover, our previous work showed that the GelSight Baby Fin Ray can dig through clutter, and classify in-shell nuts. To further examine the potential of the GelSight Baby Fin Ray, we leverage learning to distinguish nut-in-shell textures and to perform force and position estimation. We implement ablation studies with popular neural network structures, including ResNet50, GoogLeNet, and 3- and 5-layer convolutional neural network (CNN) structures.  We conclude that machine learning is a promising technique to extract useful information from high-resolution tactile images and empower soft robotics to better understand and interact with the environments.
    
\end{abstract}

\keywords{Tactile sensors, Fin Ray, contact sensing, object perception} 


\section{Introduction}

The Fin Ray finger is inspired by fish fin deformation, where the fin bends against and complies to the force applied on its structure \cite{pfaff2011application}. As a result, the Fin Ray can be implemented for different robotic tasks and, due to its simple structure, is easy to print and prototype. Although much research has focused on mechanical behavior optimization \cite{shan2020modeling, deng2021learning}, there has been sparse work on integrating sensors into Fin Rays. Fin Rays could benefit from having tactile feedback, and several works have focused on adding force sensors or using embedded cameras and machine learning to do force estimation \cite{yang20213d, xu2021compliant}. 

More recently, work has been done to integrate classically rigid camera-based sensors into various soft finger designs \cite{amini2020uncertainty, she2020exoskeleton, faris2023proprioception, 2023GelSightEA}. Specifically, camera-based sensors can provide higher-resolution details, discern object geometries, and measure contact forces \cite{yuan2017gelsight}. The original GelSight Fin Ray was able to have the passive adaptability of its compliant structure and the high-resolution tactile sensing enabled by camera sensors \cite{og_finray}. These capabilities allowed it to grasp various objects, track surface forces, and perform 3D reconstruction of the tactile deformation. Further mechanical improvements resulted in the GelSight Baby Fin Ray, which is more compliant, robust, and compact than its predecessor \cite{bbfinray}.

\begin{figure}[ht!]
	\centering
	\includegraphics[width=.5 \linewidth]{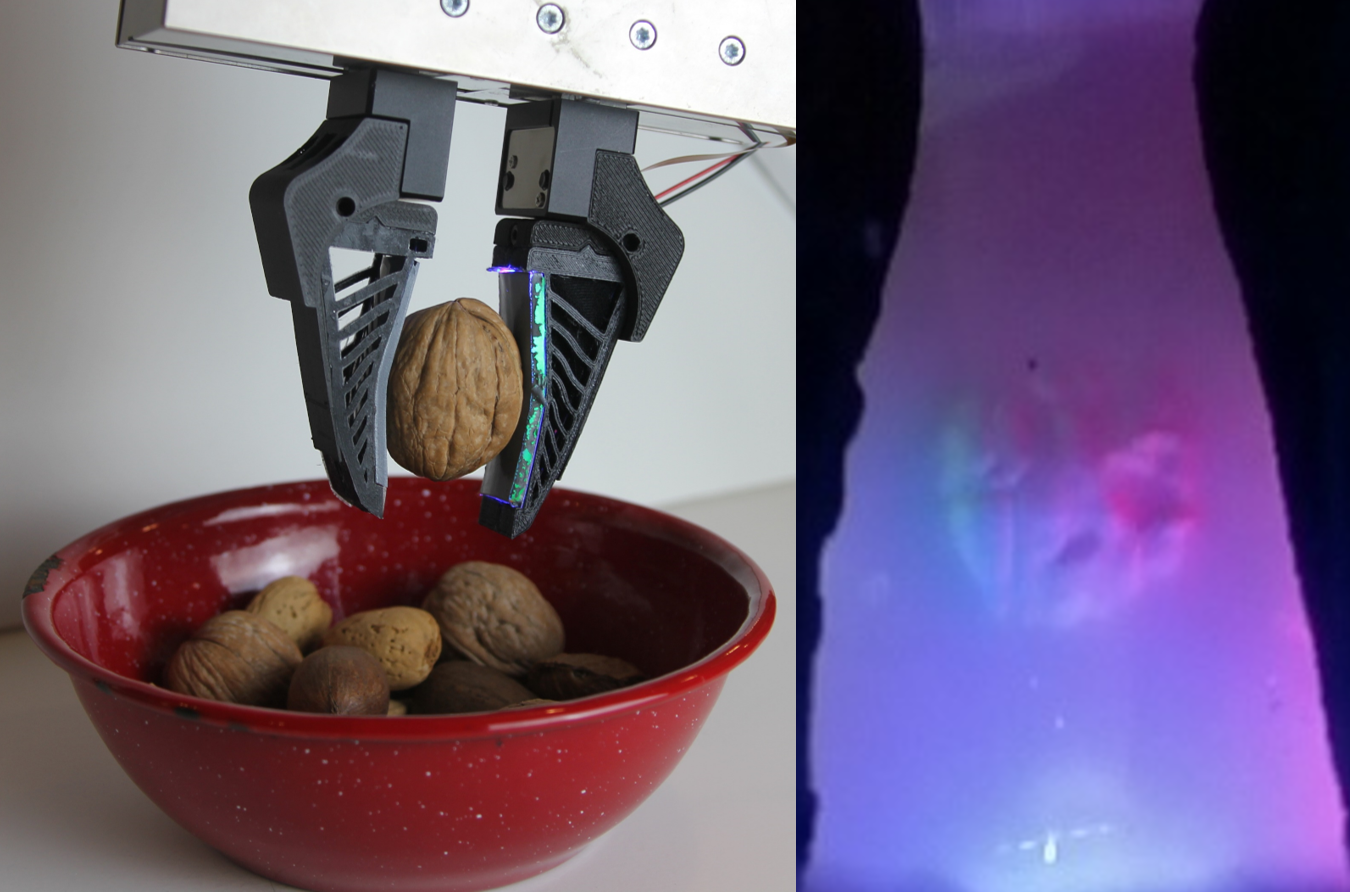}
    \vspace{-0pt}
	\caption{On the left, the GelSight Baby Fin Ray grabs a walnut; on the right is the corresponding raw tactile image \cite{bbfinray}.}
	\vspace{-8pt}
	\label{fig:teaser}
\end{figure}

 We present further work on leveraging the camera-based tactile sensor in our Fin Ray structure. In this work, we implement multiple machine learning algorithms to classify textures of nuts-in-shells and do force estimation and tactile position determination along the length of the tactile sensor. 

\section{Related Work}
Machine learning techniques have been used for multiple rigid finger sensors incorporated with camera-based tactile sensors. Guo et al. and Huang et al. are able to supplement these sensors with machine learning methods to estimate properties of solid particles and liquids inside containers \cite{guo2023estimating, huang2022understanding}. Other papers have used various learning methods to extract information about surface forces and 3D geometrical reconstruction of tactile deformation \cite{lin2022dtact, wedge, gelslim}. Still, others have done texture classification \cite{amini2020uncertainty}. On the other hand, only a few papers have performed proprioception on soft robots, i.e. state estimation or force prediction, and have not really focused on identifying textures \cite{wang2020real, scharff2021sensing, faris2023proprioception}. This work implements both object classification using texture and force estimation with a single camera sensor incorporated into a soft, compliant robot finger.

\section{Hardware}

\begin{figure}[ht]
	\centering
	\includegraphics[width= 1\linewidth]{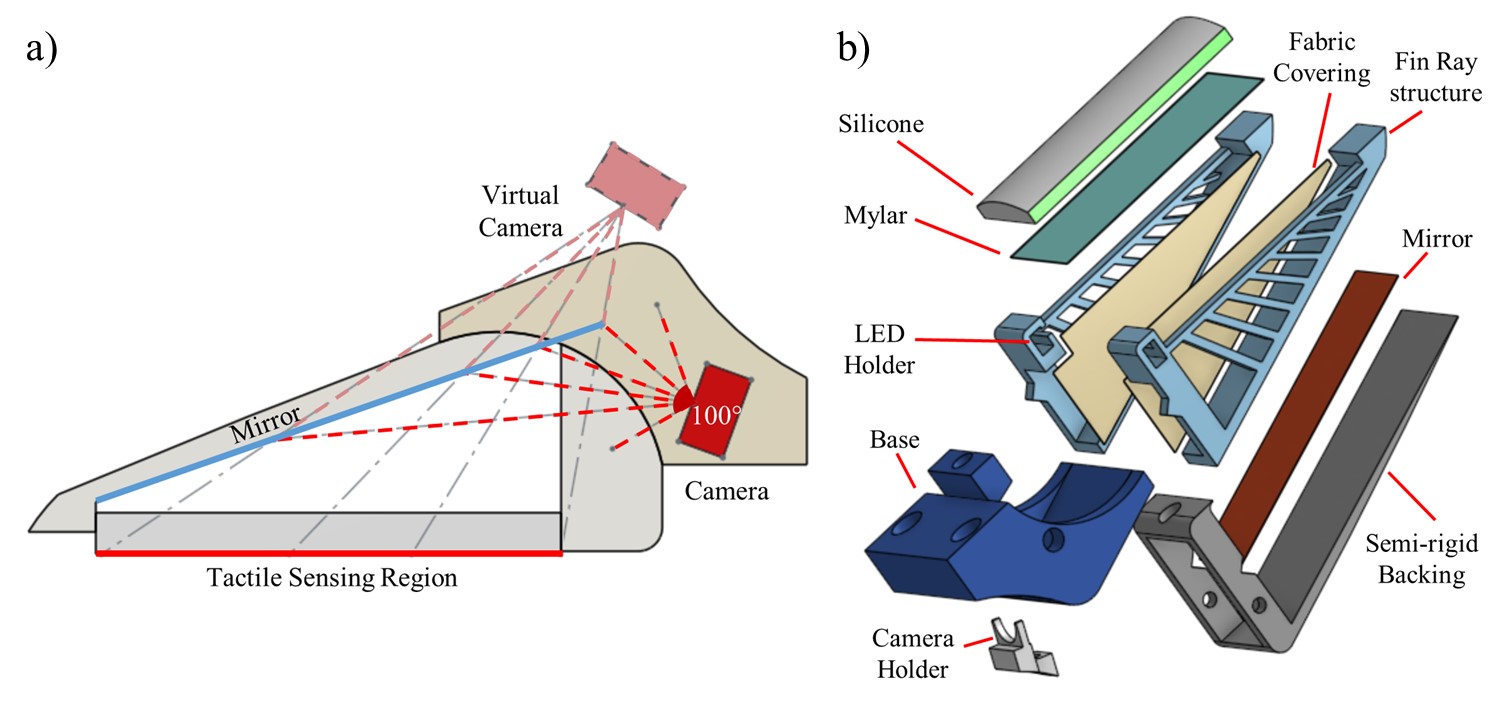}
    \vspace{-10pt}
	\caption{a) A 2D representation of the GelSight Baby Fin Ray, which shows that the camera and mirror are able to view the entire tactile sensing region. b) Exploded view of the assembly without the camera, LEDs, and diffuser \cite{bbfinray}.}
	\vspace{-10pt}
	\label{fig:explosion}
\end{figure}

We utilize the hardware design from our previous work on the GelSight Baby Fin Ray \cite{bbfinray}. A representation of this design is shown in Fig. \ref{fig:explosion}

\section{Experiment \& Results}
\subsection{Nut Classification}
One of the original tasks performed by the GelSight Baby Fin Ray paper was nuts-in-shell classification. We perform ablation studies against the ResNet50 architecture \cite{resnet} that was used in \cite{bbfinray} and compare the results with GoogLeNet \cite{googlenet}, a 3-layer and a 5-layer CNN \cite{cnn}, a support vector machine (SVM) using both radial basis functions (RBF) and polynomial functions \cite{svm}, and a k-nearest neighbors (KNN) classification algorithm \cite{knn}. For the Resnet, CNN, and GoogLeNet architectures, we use an unwarped, augmented image as the input, effectively getting rid of the ``negative'' space in the image. For both the KNN and SVM algorithms, we create a feature vector using the raw pixels in the image as the input. 

We collect data using images streamed from the raspberry pi camera via mjpg\_streamer. Nuts are pressed along the entire tactile surface at various angles. A total of 500 images per nut type are captured. The data is split into 80\% training and 20\% validation.

\begin{figure}[ht]
	\centering
	\includegraphics[width=1.0 \linewidth]{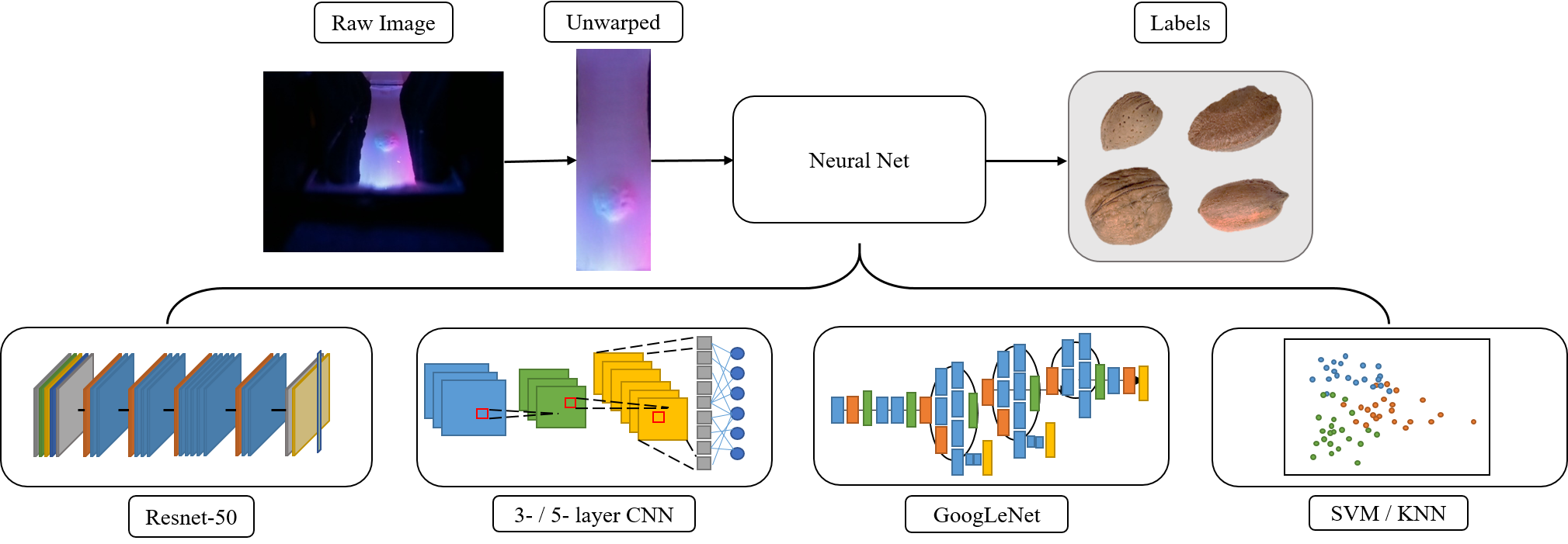}
    \vspace{-15pt}
	\caption{To classify our four classes, Almond, Brazil Nut, Pecan, and Walnut, we take the raw image, unwarp it and train it using multiple neural network architectures including ResNet50, GoogLeNet, 3- and 5-layer CNN, SVM, and KNN.}
	\vspace{-5pt}
	\label{fig:resnet}
\end{figure}

\begin{table}[!htb]
    \centering
    \caption{Accuracy of Nut Classification}
    \begin{tabular}{cccccc} \hline
    Network & Overall & Almonds & Brazil Nuts & Pecans & Walnuts \\ \hline
    3-layer CNN & 73.0\% & 73.6 \%  & 84.0 \% & 70.8 \% & 64.0 \% \\
    5-layer CNN & 87.5\% & 82.1 \%  & 87.2 \% & 91.7 \% & 89.4 \% \\
    ResNet & 95.6\% & \textbf{98.1\%} & 96.8 \% & \textbf{96.9 \%} & 91.3 \% \\
    GoogLeNet & \textbf{97.5\%} & 96.2 \%  & \textbf{98.9\%} & \textbf{96.9 \%} & \textbf{98.1\%} \\
    SVM (poly) & 86.3\% & 83.0 \%  & 86.0 \% & 79.0 \% & 97.0 \% \\
    SVM (rbf) & 79.0\% & 81.0 \%  & 77.0 \% & 68.0 \% & 90.0 \% \\
    KNN & 73.3\% & 71.0 \%  & 78.0 \% & 72.0 \% & 72.0 \% \\
    \hline \label{tab:nuts} \end{tabular}
\end{table}

Our results are shown in Table \ref{tab:nuts}. In general, the deeper networks performed better on the dataset. We believe that these results could be because deeper architectures are able to recognize more minute and complex details that might have been lost by a more shallow net or by using SVM and KNN methods. Nonetheless, all of the algorithms and neural nets performed well, indicating that the high-resolution tactile details of the nut shells are easily discernible. 

\subsection{Force Estimation}
\begin{figure}[ht]
	\centering
	\includegraphics[width=0.7 \linewidth]{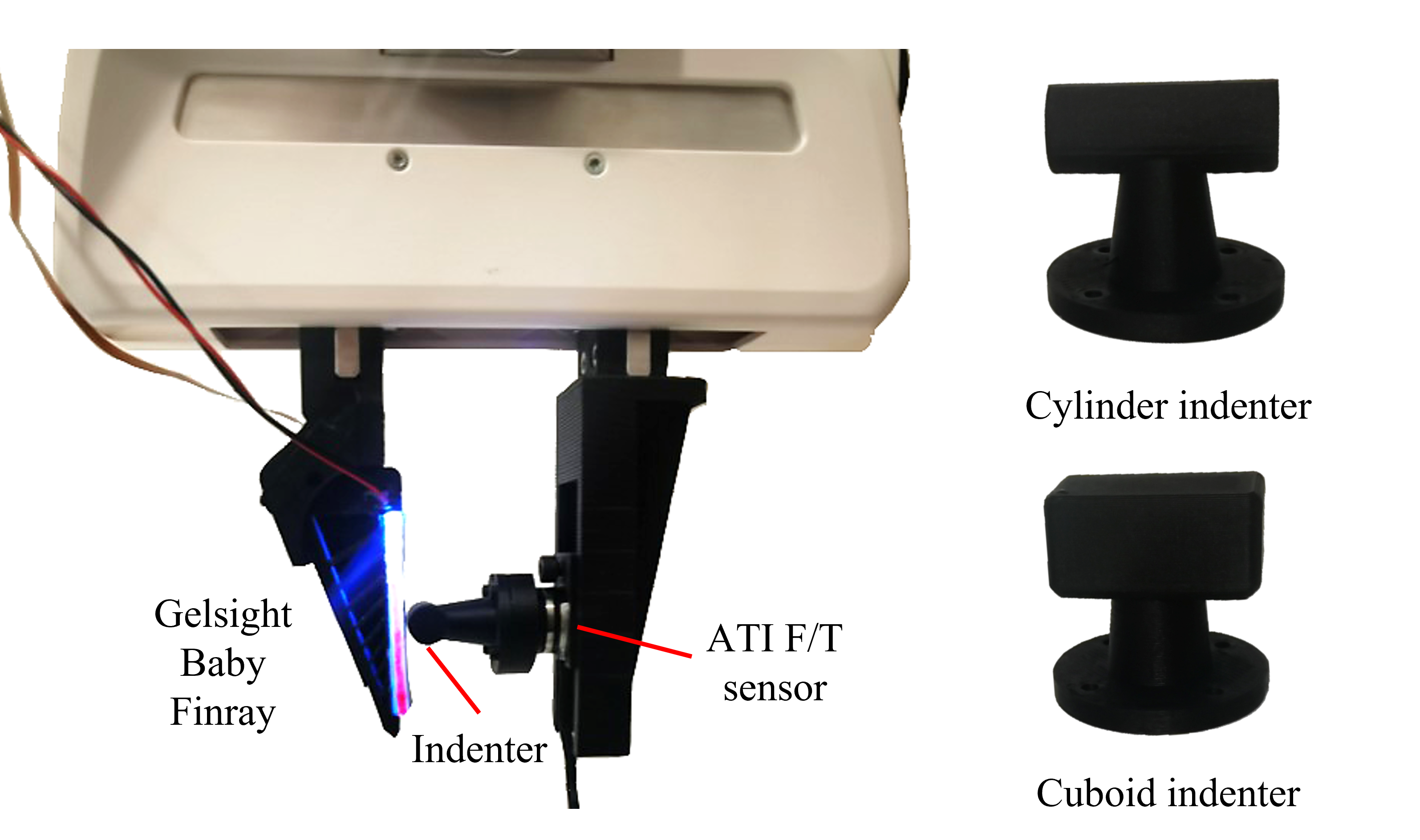}
    \vspace{-15pt}
	\caption{Experimental setup of force measuring with a cylinder indenter and a cuboid indenter.}
	\vspace{-5pt}
	\label{fig:force}
\end{figure}

We also establish learning-based force estimation with the Gelsight Baby Fin Ray. A cylinder or cuboid indenter and a Gelsight Baby Fin Ray are installed on a Franka Panda Hand, as shown in Fig. \ref{fig:force}. We program the robot hand to push indenters into the Gelsight Baby Fin Ray, while using a Nano 17 Force/Torque (ATI) sensor to record the indentation force. Concurrently, tactile images are collected via mjpg\_streamer. Contacts are made at 10 mm to 50 mm along the Fin Ray length, and the range of normal force is from 0 to 25 N. A total of 60000 images per indenter type are captured. The network predicts both normal force and contact position.

\begin{table}[!htb]
    \centering
    \caption{Estimation Error of Contact Position and Normal Force Estimation}
    \begin{tabular}{ccc} \hline
    Network & Contact Position Error (mm) & Normal Force Error (N) \\
    3-layer CNN &  2.38 & 1.24 \\
    5-layer CNN &  5.09  & 2.21  \\
    ResNet50 &     2.15  & 1.21 \\
    GoogLeNet &    \textbf{2.02}  & \textbf{1.18}  \\ \hline \label{tab:force} \end{tabular}
\end{table}


We perform ablation studies with the ResNet50, GoogLeNet, 3- and 5-layer CNNs. The 5-layer CNN tends to have large gradients in this regression task, resulting in the worst accuracy. Similar to the classification task, GoogLeNet outperforms all the other architectures in both contact position estimation and normal force estimation. The reason might be that the GoogleNet inception modules can effectively analyze the contact imprints with different sizes.

\section{Conclusion}
Camera-based tactile sensors combined with learning methods are useful for simultaneously determining texture identification, force estimation, and position of tactile displacement. Overall, these abilities allow us to minimize the amount of electronics that one finger needs and to design stratagems that allow us to integrate these sensors into a compliant structure, like the Fin Ray. To conclude, these soft tactile sensors have potential uses for many manipulation tasks in the future.

\clearpage

\acknowledgments{Toyota Research Institute, the Office of Naval Research, and the SINTEF BIFROST (RCN313870) project provided funds to support this work.}


\bibliography{ref}  

\end{document}